\documentclass[10pt,twocolumn,letterpaper]{article}

\usepackage{wacv}
\usepackage{times}
\usepackage{epsfig}
\usepackage{graphicx}
\usepackage{amsmath}
\usepackage{amssymb}
\usepackage{booktabs}
\usepackage{mathtools}
\usepackage{adjustbox}
\usepackage{array}
\usepackage{multirow}
\usepackage{enumitem}
\usepackage{colortbl}
\usepackage{bm}
\usepackage{lipsum}
\usepackage{indentfirst}
\usepackage{mathrsfs}
\usepackage[T1]{fontenc}
\usepackage{bm}
\usepackage{bbding}
\usepackage[section]{placeins}
\usepackage{float}
\usepackage{caption}

\usepackage[accsupp]{axessibility}

\def\wacvPaperID{268} % Enter the WACV Paper ID here

\wacvfinalcopy % *** Uncomment this line for the final submission

\ifwacvfinal
\usepackage[breaklinks=true,bookmarks=false]{hyperref}
\else
\usepackage[pagebackref=true,breaklinks=true,colorlinks,bookmarks=false]{hyperref}
\fi

\usepackage[capitalize]{cleveref}
\crefname{section}{Sec.}{Secs.}
\Crefname{section}{Section}{Sections}
\Crefname{table}{Table}{Tables}
\crefname{table}{Tab.}{Tabs.}

\begin{document}

%%%%%%%%% TITLE
\title{Kernel-Aware Burst Blind Super-Resolution}

\author{Wenyi Lian\\
Department of Information Technology\\
Uppsala University, Sweden\\
{\tt\small wenyi.lian.7322@student.uu.se}
% For a paper whose authors are all at the same institution,
% omit the following lines up until the closing ``}''.
% Additional authors and addresses can be added with ``\and'',
% just like the second author.
% To save space, use either the email address or home page, not both
\and
Shanglian Peng\\
School of Computer Science\\
Chengdu University of Information Technology, China\\
{\tt\small psl@cuit.edu.cn}
}

\maketitle
\thispagestyle{empty}

%%%%%%%%% ABSTRACT
\begin{abstract}
% We address the problem of reconstructing a high-resolution (HR) image from low-quality burst image sequence acquired from modern handheld devices.
Burst super-resolution (SR) technique provides a possibility of restoring rich details from low-quality images. However, since real world low-resolution (LR) images in practical applications have multiple complicated and unknown degradations, existing non-blind (e.g., bicubic) designed networks usually suffer severe performance drop in recovering high-resolution (HR) images. In this paper, we address the problem of reconstructing HR images from raw burst sequences acquired from a modern handheld device. The central idea is a kernel-guided strategy which can solve the burst SR problem with two steps: kernel estimation and HR image restoration. The former estimates burst kernels from raw inputs, while the latter predicts the super-resolved image based on the estimated kernels. Furthermore, we introduce a pyramid kernel-aware deformable alignment module which can effectively align the raw images with consideration of the blurry priors. Extensive experiments on synthetic and real-world datasets demonstrate that the proposed method can perform favorable state-of-the-art performance in the burst SR problem. Our codes are available at \url{https://github.com/shermanlian/KBNet}.

\end{abstract}

%%%%%%%%% BODY TEXT
\section{Introduction}

With the growing popularity of built-in smartphone cameras, the multi-frame super-resolution (MFSR) has drawn much attention due to its high practical potential to recover rich details from several low-quality images \cite{bhat2021deep,bhat2021deepre,lecouat2021lucas}. Compared with single image super-resolution (SISR), MFSR can provide complementary information from subpixel shifts, avoiding aliasing artifacts and losing details \cite{park2003super,nasrollahi2014super}. Typically, we explicitly model the multi-frame degradation process as:
\begin{equation}
    \centering
    \bm{x}_i = (\bm{k}_i \otimes {\cal T}_i\bm{y})_{\downarrow_s} + \eta_i,
    \label{eq:mf_degrad}
\end{equation}
where $\bm{y}$ is the original HR image, $\{\bm{x}_i\}_1^N$ is the observed low-resolution image bursts. $\bm{k}_i$ and ${\cal T}_i$ denote the blur kernel and scene motion transform, respectively. $\otimes$ represents convolution operation and ${\downarrow_s}$ is the subsequent downsampling with scale factor $s$. $\eta_i$ is an white Gaussian noise that is independent to LR images.

% \begin{figure}[t]
%     % \setlength{\abovecaptionskip}{0.05in}
%     % \setlength{\belowcaptionskip}{-0.05in}
%   \centering
%   \includegraphics[width=1.\linewidth]{figs/new_burst_kernel.pdf}
%   \caption{An example of estimated blur kernels from a real-world burst sequence.}
%   \label{fig:burst_kernels}
% \end{figure}

\begin{figure}[t]
    \setlength{\abovecaptionskip}{0.05in}
  \centering
  \includegraphics[width=1.\linewidth]{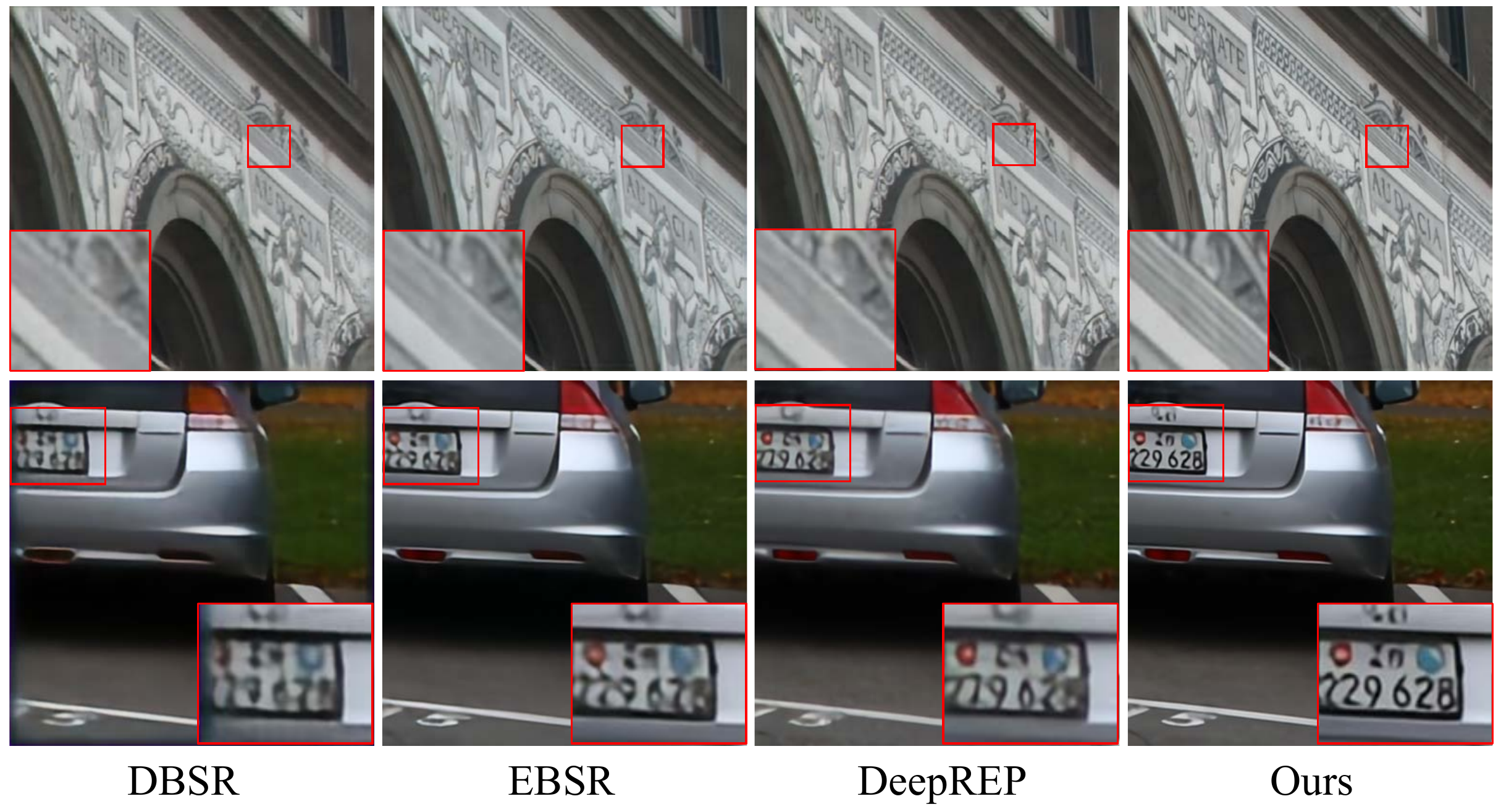}
  \caption{We propose a degradation guided framework to handle the burst super-resolution problem on both real-world dataset (top row) and synthetic dataset (bottom tow). The proposed method outperforms existing state-of-the-art MFSR approaches DBSR \cite{bhat2021deep}, EBSR \cite{luo2021ebsr} and DeepREP \cite{bhat2021deepre}. Our method is effective in restoring edges and details.}
  \label{fig:teaser}
\end{figure}

Most existing MFSR methods assume that the blur kernels are known (e.g., bicubic) and the same for all frames \cite{bhat2021deep,bhat2021deepre,lecouat2021lucas,luo2021ebsr}. Under this assumption, these MFSR methods can achieve dramatic performance to search for the best inverse solution for the bicubicly downsampling degradation. However, they often suffer severe performance drop when applied to real-world applications of that the kernel is actually derived from cameras' intrinsic parameters that are complicated and inconsistent for all burst frames \cite{efrat2013accurate,gu2019blind,yang2014single}. Moreover, multiple blurry inputs would make the restoration difficult and lose details (See Fig. \ref{fig:teaser}). And transferring the bicubicly designed model to unknown degradation images is also inefficient. To this end, we pay more attention to the model that tackles degradation of multiple unknown blur kernels, \textit{i.e.} burst blind SR.

Single image blind SR has been well studied in recent works \cite{zhang2018learning,xu2020unified,gu2019blind,bell2019blind,luo2020unfolding}, which often need to sequentially estimate blur kernel (or its embedding), and then restore the SR image based on that kernel. However, the overall optimization of blind SR is usually alternating, complicated, and time-consuming \cite{gu2019blind,hui2021learning}. Such a problem could be even more serious when facing burst blind SR, where each frame has a specific blur kernel and an irregular motion displacement. So far little work has focused on the blind property of the burst SR. The common solution is to train a deep model directly on the bicubicly designed synthetic dataset, then finetune it on another real-world dataset \cite{bhat2021deep,dudhane2021burst,luo2021ebsr,lecouat2021lucas,Umer_2021_ML4PS}. However, it is quite challenging to set up a real-world dataset due to the degradation is always sensor-specific and the images captured by smartphones (LR) and DSLR cameras (HR) usually have different qualities and image signal processors (ISP).

In this paper, we address above issues by proposing a kernel-aware raw burst SR method based on the multi-frame degradation model (\cref{eq:mf_degrad}), called \textit{KBNet}, which takes into account the inconsistencies of blur kernels between frames and can learn a practical multi-frame super-resolver using synthetic data. The KBNet consists of two neural networks: a kernel modeling network that estimates blur kernels for each burst frames, and a restoring network that predicts the super-resolved image by fusing the information of all frames and the corresponding estimated kernels. To make full use of the degradation information from kernels, the restoring network employs an Adaptive Kernel-Aware Block (AKAB) to extract deblurred clean features and a Pyramid Kernel-Aware Deformable convolution ( Pyramid KAD) module to align and fuse multiple complementary features. Our contributions are summarized as follows:
\begin{itemize}
    \item We consider the inconsistent degradation of different frames and propose a novel kernel-aware network, named \textit{KBNet}, for raw burst image blind super-resolution, which makes a substantial step towards developing read-world practical MFSR applications.
    \item We propose a degradation-based restoring network that uses adaptive kernel-aware blocks and a pyramid kernel-aware deformable alignment module to restore SR image based on blur kernels.
    \item Extensive experiments demonstrate that the proposed method achieves an art performance on various synthetic datasets and real images. 
\end{itemize}
\section{Related work}

\subsection{Single Image Super-Resolution}
% Numerous works have 
SISR is a problem of trying to recover a high-resolution image from its degraded low-resolution version. In the past few years, numerous works based on the neural network have achieved tremendous performance gains over the traditional SR methods \cite{dong2014learning,dong2015image,hui2018fast,kim2016accurate,ledig2017photo,wang2018esrgan,zhang2018residual,lim2017enhanced,lian2022sliding}. 
Since the pioneering work SRCNN \cite{dong2014learning}, most subsequent works focus on optimizing the network architecture \cite{dong2015image,lai2017deep,zhang2018residual,lim2017enhanced,dai2019second} and loss function \cite{johnson2016perceptual,ledig2017photo,wang2018esrgan,lugmayr2020srflow}. These methods are difficult to apply to real-world applications due to their ill-posed nature.
% and domain gap between real and synthesized data.

\begin{figure*}[t]
\centering
   \includegraphics[width=.95\linewidth]{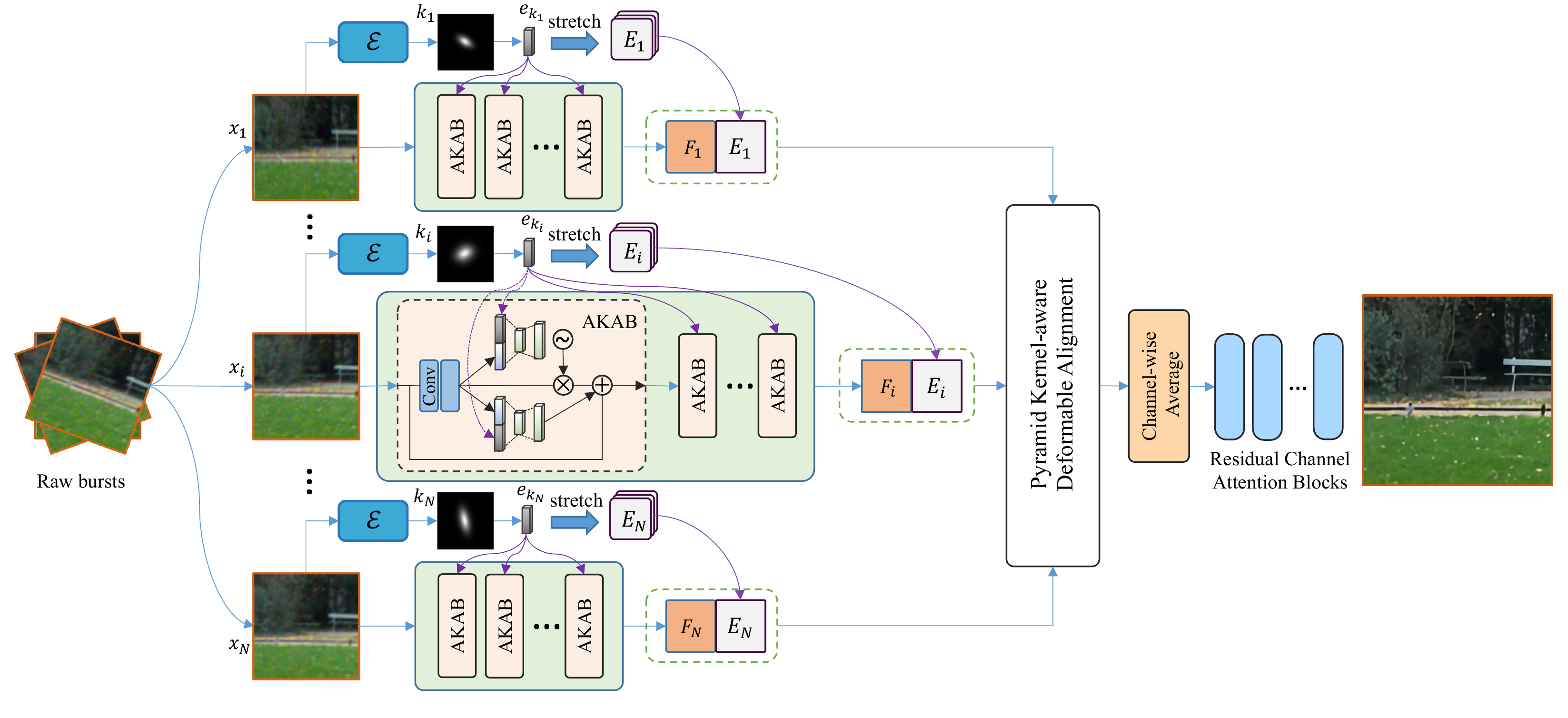}
   \caption{The overview of our method. The inputs are a set of RAW burst images $\{x_i\}_{i=1}^N$. We predict blur kernel for each frame through a simple CNN network, as estimator $\cal E$. The estimated kernels are reduced to embeddings by PCA and fed into groups of adaptive kernel-aware blocks (AKABs) to extract clean features. And we also stretch the kernel embeddings to degradation maps so as to concatenate them with clean features. These features are then aligned with the reference frame through a pyramid kernel-aware deformable alignment module. We fuse these aligned features with a channel-wise averaging strategy and use residual channel attention blocks to reconstruct the SR result.}
   \label{fig:ov}
\end{figure*}

\subsection{Multi-Frame Super-Resolution}

MFSR is an active area and has been well studied in the last three decades. Tsai and Huang \cite{tsai1984multiframe} are the first that propose to tackle the MFSR problem. They propose to restore HR image in the frequency domain with known translations between frames. Peleg et al. \cite{peleg1987improving} and Irani and Peleg \cite{irani1991improving} propose an iterative back-projection approach that can sequentially estimate HR image and synthesizes LR image. Later works \cite{bascle1996motion,elad1997restoration,haris2019recurrent,schultz1996extraction,hardie1998high} improve this method with a maximum a posteriori (MAP) model and a regularization term. Takeda et al. \cite{takeda2006robust,takeda2007kernel} introduce a kernel regression technique for super-resolution and Wronski et al. \cite{wronski2019handheld} applies it on fusing aligned input frames. 

Recently, several works \cite{deudon2020highres,kawulok2019deep,molini2019deepsum} propose to incorporate deep learning to handle the MFSR problem in remote sensing applications. Bhat et al. \cite{bhat2021deep,bhat2021ntire} introduce a real-world dataset and propose an attention-based based fusion approach for MFSR. And they further improve the model to handle both SR and denoising by transforming the MAP framework to a deep feature space. Luo et al. \cite{luo2021ebsr,luo2022bsrt} introduce the deformable convolution to MFSR and show its effectiveness of handling the alignment between frames. Bruno et al. \cite{lecouat2021lucas} propose an effective hybrid algorithm building on the insight from \cite{wronski2019handheld}. Akshay et al. \cite{dudhane2021burst} propose to create a set of pseudo-burst features that makes it easier to learn distinctive information of all frames.

\subsection{Blind Super-Resolution}

Blind SR assumes that the blur kernels of degradation are unavailable. In recent years, the blind SR problem has drawn much research attention since it is close to real-world scenarios \cite{michaeli2013nonparametric}. Zhang et al. \cite{zhang2018learning} firstly propose to extract principal components of the Gaussian blur kernels and stretch and concatenate them with LRs to get degradation-aware SR images. Subsequently, Gu et al. \cite{gu2019blind} modify the strategy in \cite{zhang2018learning} by concatenating kernel embeddings with deep features. Luo et al. \cite{luo2020unfolding} and Zhang et al. \cite{zhang2020deep} propose to unfold the blind SR problem as a sequential two-step solution which can be alternately optimized. Hussein et al. \cite{hussein2020correction} propose a closed-form correction filter to transform blurry LR images to bicubicly downsampled LR images. Luo et al.~\cite{luo2022deep} reformulate the kernel to LR space and then they apply a deconvolution to get deblurred SR images. Moreover, ZSSR \cite{shocher2018zero,bell2019blind} and MSSR \cite{soh2020meta} also can be applied to blind SR, where the training is conducted as test time, and it can exploit the internal information of the LR by using an image-specific degradation.
\section{Method}

This section describes the main techniques of the proposed KBNet for raw burst blind super-resolution. As shown in \cref{fig:ov}, we first estimate the blur kernel for each frame and obtain its embedding vector through the principal component analysis (PCA). By taking the LR frame and the corresponding degradation kernel embedding as inputs, we can extract clean features with several adaptive kernel-aware blocks (AKAB). Then we stretch the kernel to degradation maps so that the feature and the kernel embedding can be concatenated and sent into the Pyramid kernel-aware deformable (KAD) alignment module. After that, we fuse these aligned clean features by a channel-wise average operation and then restore the HR image like a traditional SISR model through several residual channel attention blocks (RCAB)~\cite{zhang2018image}.

\subsection{Problem Formulation}
Given raw burst images $\{x_i\}_{i=1}^N, x_i \in \mathbb{R}^{h\times w \times 1}$ captured from the same scene and the scale factor $s$, our goal is to extract and fuse the complementary information between bursts and restore a high-quality image $y \in \mathbb{R}^{sh\times sw \times 3}$ with rich details. In our scenario, each input $x_i$ is a one-channel raw image, while the output is usually a RGB image. The degradation of burst SR is model as \cref{eq:mf_degrad}. We assume that the degradation adopts anisotropic Gaussian kernels and the noise $\eta_i$ is independent to the LR image $x_i$. The burst blind SR problem can be tackled by solving the following maximum a posteriori (MAP) problem:
\begin{equation}
    \centering
    \arg\min_{k, y} \sum_{i=1}^N || \bm{x}_i - (\bm{k}_i \otimes {\cal T}_i\bm{y})_{\downarrow_s} ||_2^2 + \phi ({\bm y}) + \psi({\bm k}_i),
    \label{eq:map}
\end{equation}
where $\phi ({\bm y})$ and $\psi({\bm k})$ are the parameterized prior regularizers. Since all of the kernels of multiple frames are unknown variables, the overall problem is extremely difficult and challenging. Inspired by recent success of single image blind SR \cite{gu2019blind,luo2020unfolding}, we decompose this problem into two sequential steps:
\begin{equation}
\centering
    \begin{cases}
    {\bm k}_i = {\cal E}({\bm x}_i; \theta_e)\\
    {\bm y} = {\cal R}(\{{\bm x}_i, {\bm k}_i\}_{i=1}^N; \theta_r),
\end{cases}
\label{eq:two-step}
\end{equation}
where ${\cal E}(\cdot)$ denotes the kernel estimator that predicts kernels for each frame of the raw bursts and ${\cal R}(\cdot)$ denotes the restorer that restores HR image based on LR frames and the estimated kernels. $\theta_{e}$ and $\theta_r$ are the parameters of the estimator and restorer, respectively.

\subsection{Kernel Estimator}
\begin{figure}[t]
\setlength{\abovecaptionskip}{0.05in}
\setlength{\belowcaptionskip}{-0.1in}
  \centering
  \includegraphics[width=.95\linewidth]{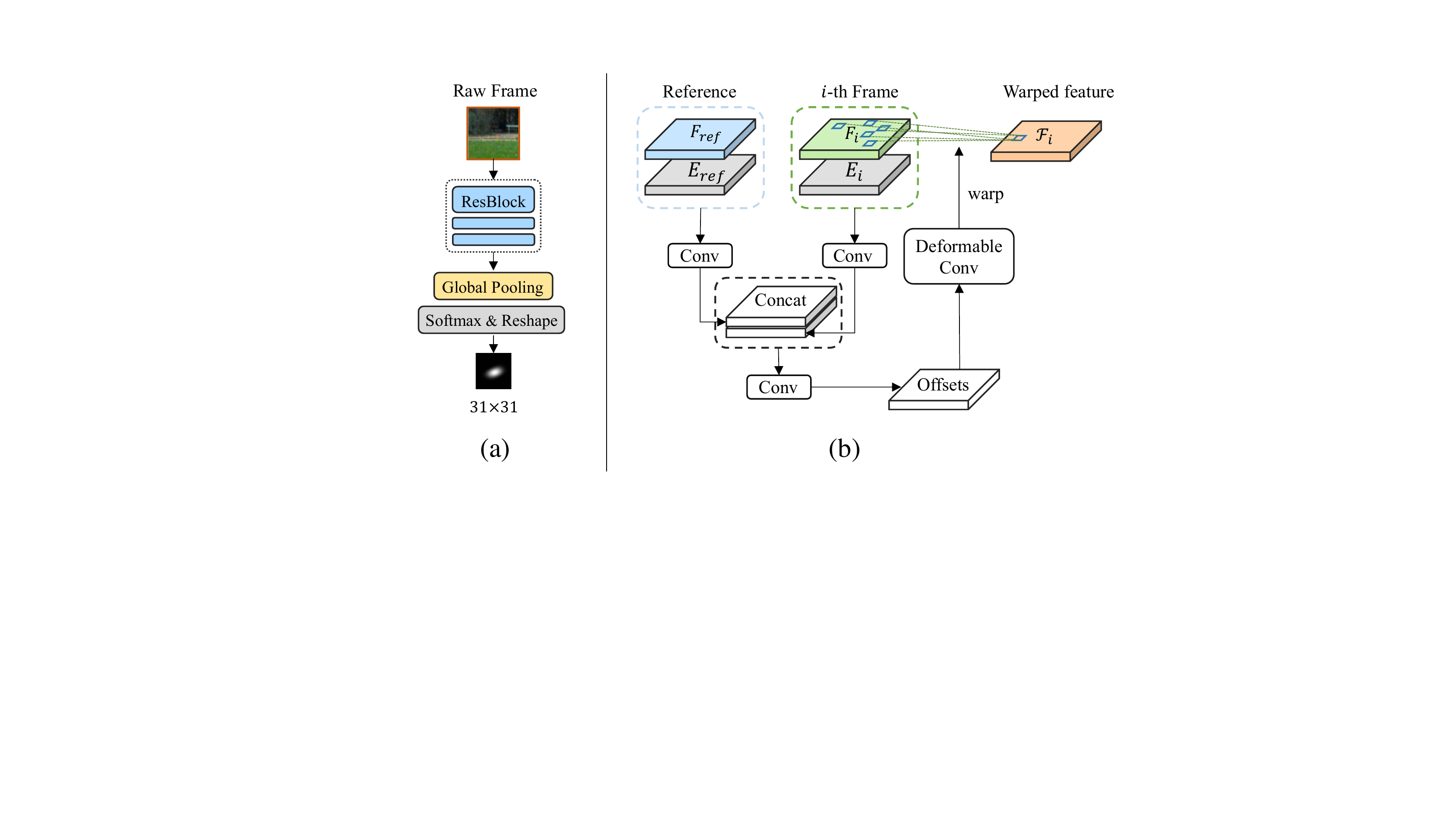}
  \caption{(a) Network architecture of the kernel estimator. (b) Kernel-aware deformable (KAD) alignment module.}
  \label{fig:estk}
\end{figure}

In order to obtain the degradation kernel and help the SR model produce visual pleasant images, we introduce an estimator $\cal E$ to predict blur kernels for all frames. The network architecture of the estimator is illustrated in \cref{fig:estk}(a). It consists of three simple steps: feature extraction, global average pooling, and reshape operation. Note that as a widely used kernel prior, we use the softmax function in the last layer so that the kernel could be sum to one. Moreover, we use ground truth kernels as strong supervision to optimize the estimator network. The objective is to minimize the ${\cal L}_1$ loss between the estimated kernel and the ground truth as
\begin{equation}
    \centering
    \theta_{e} = \arg\min_{\theta_{e}} \sum_{i=1}^N || \bm{k}_i - {\cal E}(\bm{x}_i;\theta_{e}) ||_1.
    \label{eq:k_loss}
\end{equation}
In practice, the body of the estimator consists 3 residual blocks. The overall network is simple, lightweight yet efficient. Since the estimated kernels are sent to the restorer to help to super-resolve images. We can jointly optimize \cref{eq:k_loss} with the restorer model together to construct an end-to-end blind burst SR training scheme.

\subsection{Adaptive Kernel-Aware Block (AKAB)}

Most blind SR methods \cite{zhang2018learning,gu2019blind,luo2020unfolding} tend to stretch the kernel embedding to a full-sized degradation map $E_i$ and concatenate it with deep features to extract useful information to predict SR image. However, such a strategy is usually inefficient and computationally costly \cite{hui2021learning}.
In this work, we propose the adaptive kernel-aware block (AKAB) that can utilize low-dimensional embedding and statistical information (such as the mean value of each feature) to extract deep informative features. As illustrated in the center of \cref{fig:ov}, the LR feature is firstly sent to two convolution layers, and then squeezed to one-dimensional embedding by a global average pooling layer. After that, the feature embedding $e_{{\bm x}_i}$ is concatenated with the corresponding kernel embedding $e_{{\bm k}_i}$ to perform a residual affine attention mechanism which is defined as:
\begin{equation}
\centering
    {\bm x}_i^{out} = {\gamma}(e_{{\bm x}_i}, e_{{\bm k}_i}) \odot {\bm x}_i + {\beta}(e_{{\bm x}_i}, e_{{\bm k}_i}) + {\bm x}_i,
\label{eq:aa}
\end{equation}
where $\gamma(\cdot)$ and $\beta(\cdot)$ denote the scaling and shifting functions, both of which consist of two linear layers as 
\begin{gather}
    {\gamma}(e_{{\bm x}_i}, e_{{\bm k}_i}) =  g(w_2\sigma(w_1{\cal C}(e_{{\bm x}_i}, e_{{\bm k}_i})),\\[0.3em]
    {\beta}(e_{{\bm x}_i}, e_{{\bm k}_i}) =  v_2\sigma(v_1{\cal C}(e_{{\bm x}_i}, e_{{\bm k}_i})),
\label{eq:gamma}
\end{gather}
where ${\cal C}(\cdot)$ represents the concatenating operation across the channel dimension. $w_1, w_2$ and $v_1, v_2$ denote the linear layers for $\gamma$ and $\beta$, respectively. $\sigma$ represents a non-linear activation (e.g., ReLU), and $g$ denotes the sigmoid function. 
Moreover, the shifting network $\beta$ focuses on extracting and aggregating channels' information to enrich the deep features to boost the performance of feature alignment.

We compose several AKABs as a powerful feature extractor so as to obtain cleaner features that implicitly embed the degradation information of kernels. Then we could align different frames in the feature level by the following kernel-aware alignment module.

\begin{figure*}[t]
  \centering
   \includegraphics[width=.95\linewidth]{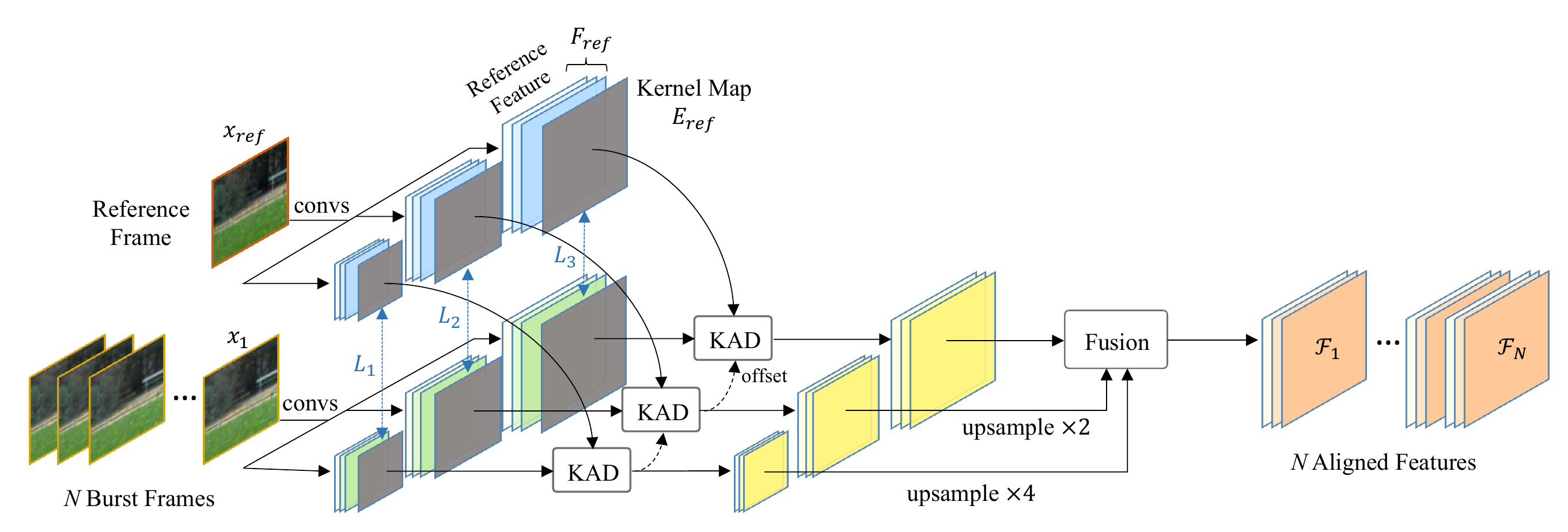}
   \caption{Pyramid kernel-award deformable alignment module. Starting from level $L_1$, we align all features with reference using KAD, and the predicted offset are sent to the next level to be concatenated with features to improve the offset estimation. Then each level's output are upsampled to have the same sizes so we can fuse them to obtain the aligned output features.}
   \label{fig:kdcn}
\end{figure*}

\subsection{Pyramid Kernel-Aware Deformable Alignment}
Deformable convolution network (DCN)~\cite{jaderberg2015spatial,dai2017deformable,zhu2019deformable} has demonstrated its effectiveness of aligning features from multiple frames \cite{tian2020tdan,chan2020understanding,wang2019edvr,chan2021basicvsr++}. 
% we introduce DCN into burst SR as the alignment module to effectively aggregate information from different frames. 
However, training the primitive DCN in the multi-degradation scenario is difficult since the different features of the same scene may have different manifestations. Thus we introduce the degradation information into the alignment process, as a kernel-aware deformable convolution (KAD) module, which could help the DCN to learn accurate offsets without being affected by various degradations.

\noindent \textbf{KAD module.} 
The overview of the KAD alignment module is shown in \cref{fig:estk} (b). Specifically, we first simply stretch the reduced kernel $e_{k_i} \in \mathbb{R}^{t}$ to degradation map $E_i \in \mathbb{R}^{t \times H \times W}$. Then given the reference feature $F_{ref}$ and the input feature $F_i$ from the $i$-th LR frame, we concatenate these features with their corresponding kernel embedding maps $E_{ref}$ and $E_i$ to predict the deformable offsets $\Delta {\bm f}_i$ as
\begin{equation}
\centering
    \Delta {\bm f}_i = {\cal O}(F_{ref}, E_{ref}, F_{i}, E_i),
\label{eq:offset}
\end{equation}
where $\cal O$ is the offset predictor. Then we can get the aligned feature by warping $F_i$ with $ \Delta {\bm f}_i$~\cite{zhu2019deformable}:
\begin{equation}
\centering
    {\cal F}_i = \mathrm{warp}(F_i, \Delta {\bm f}_i).
\label{eq:warp}
\end{equation}
% where ${\cal W}$ denotes the spatial warping operation. 

\noindent \textbf{Pyramid KAD.} 
To address large camera motions, we further propose a pyramid alignment structure on the top of the KAD module. Specifically, as shown in \cref{fig:kdcn}, we first downsample all features and kernel maps with convolution layers with strides 1, 2 and 4 to get 3 different levels of pyramid. Then we perform alignment for $N$ burst frames on each pyramid level based on the KAD. And the aligned pyramid features are scaled to be the same sizes and fused as the final aligned features. By doing so, we can effectively aggregate multi-scale information from multiple images to reconstruct SR with rich details.
Empirically, we choose the first frame as the reference.

% \begin{figure}[t]
%   \centering
%   \includegraphics[width=.6\linewidth]{figs/kdcn.pdf}
%   \caption{Network structure of the kernel-aware deformable alignment module.}
%   \label{fig:kdcn}
% \end{figure}

\subsection{Fusion and Reconstruction}

Once the features are all aligned, we can combine information across the individual frames to generate a merged feature with rich details. Unlike previous works that use attention-based weights \cite{bhat2021deep} or recursive operation \cite{deudon2020highres} in the fusion, we adopt an effective channel-wise average fusion strategy as shown in \cref{fig:ov}. There are two main advantages of averaging features: firstly, the operation is fast and allows us to work with the arbitrary number of frames in both training and inference. Secondly, since the inputs are noisy, averaging all frames can reduce additional noise as a traditional denoiser. Based on the fused feature, we can reconstruct the results with advanced SR networks. Practically, we employ residual channel attention blocks (RCABs)~\cite{zhang2018image} as the reconstruction body. The objective function of the SR reconstruction network is defined via ${\cal L}_1$ loss as:
\begin{equation}
    \centering
    \theta_{r} = \arg\min_{\theta_{r}} || \bm{y} - SR(\{\bm{x}_i\}_{i=1}^N;\theta_{r})||_1.
    \label{eq:sr_loss}
\end{equation}
To avoid the phenomenon of kernel mismatching \cite{zhang2018learning,gu2019blind}, we jointly optimize the estimator and the reconstruction module in an end-to-end manner. 

% \newpage

\section{Experiments}

% In this section, we perform a comprehensive evaluation of the proposed method on the synthetic dataset and real-world BurstSR dataset \cite{bhat2021deep}.

\begin{table*}[bt]
\setlength{\abovecaptionskip}{0.05in}
\setlength{\belowcaptionskip}{-0.05in}
\centering
\resizebox{1.\linewidth}{!}{
\begin{tabular}{lcccccccccc}
\toprule
% \hline
\multirow{2}{*}{\textbf{Method}}  & \multirow{2}{*}{\textbf{AKAB}}     & \multirow{2}{*}{\textbf{RCAB}}  & \multirow{2}{*}{\textbf{KAD}}   & \multirow{2}{*}{\textbf{Pyramid KAD}}     & \multicolumn{3}{c}{\textbf{Synthetic}}   & \multicolumn{3}{c}{\textbf{Real-World}}   \\ \cmidrule(lr){6-8} \cmidrule(lr){9-11}

    &    &      &     &      &PSNR$\uparrow$   & SSIM$\uparrow$  & LPIPS$\downarrow$  &PSNR$\uparrow$   & SSIM$\uparrow$  & LPIPS$\downarrow$    \\  \midrule 

KBNet-A    & \XSolidBrush     &\XSolidBrush   & \XSolidBrush     & \XSolidBrush  & 35.24   & 0.8927 & 0.1751     & 47.12   & 0.9797 & 0.0320 \\
KBNet-B    & \Checkmark     & \XSolidBrush       & \XSolidBrush     & \XSolidBrush  & 36.53 & 0.9067 & 0.1422   & 47.61   & 0.9818 & 0.0302    \\
KBNet-C    &\Checkmark    & \Checkmark       & \XSolidBrush     & \XSolidBrush  & 36.74  & 0.9142 & 0.1383     & 47.68   & 0.9821 & 0.0298   \\
KBNet-D    &\Checkmark     & \Checkmark       & \Checkmark    & \XSolidBrush   & 36.87  & 0.9185 & 0.1290     & 47.87   & 0.9830 & 0.0278   \\
KBNet-E    &\Checkmark     & \Checkmark       & \Checkmark    & \Checkmark   & 37.29  & 0.9219 & 0.1203     & 48.27   & 0.9856 & 0.0248   \\

% \hline
\bottomrule
\end{tabular}}
% }
\caption{Ablation study on our main components. The baseline is a multi-frame SR network that adopts normal deformable convolution to align frames and reconstructs SR results by several residual blocks.}
\label{table:ablation}
\end{table*}

\subsection{Datasets and Implementations}
\label{ex:setting}

\noindent \textbf{Synthetic dataset.}
Our method is trained on Zurich RAW to the RGB dataset \cite{ignatov2020replacing} which consists of 46,839 HR images. For the synthetic setting, we focus on the anisotropic Gaussian kernels. Following \cite{bell2019blind}, we fix the blur kernel size to 31, The kernel width of both axes are uniformly sampled in the range $[0.6, 5]$. And we also rotate the kernel by an angle uniformly distributed in $[-\pi,\pi]$. The RAW burst images are synthesized by randomly translating and rotating a high-quality sRGB image, and blurring and downsampling it with kernels generated from the above procedure as \cref{eq:mf_degrad}. In the RAW space, we add noises draw from Poisson-Gaussian distribution with sigma 0.26. Then we convert the low-quality images to RAW format using an inverse camera pipeline \cite{brooks2019unprocessing}. The test sets are generated by applying anisotropic Gaussian kernels on 1204 HR images of the validation set in \cite{ignatov2020replacing} with different kernel width ranges of $[0.0, 1.6]$, $[1.6, 3.2]$ and $[3.2, 4.8]$. We assign different random seed values to ensure that different blur kernels are selected for different images. PSNR, SSIM \cite{wang2004image}, and the learned perceptual score LPIPS \cite{zhang2018unreasonable} are used as the evaluation metrics on synthetic datasets.

\noindent \textbf{Real-world dataset.}
For real-world image evaluation, we use the BurstSR dataset~\cite{bhat2021deep} which contains pairs of real-world burst images and corresponding ground truth HR images captured by a handheld smartphone camera and DSLR camera, respectively. Each burst in this dataset contains 14 raw images and is cropped to $160\times160$.
% To illustrate the capability of our method for real-world images, \textit{we only take the validation dataset with 882 raw bursts for testing}. 
We perform super-resolution by a scale factor of 4 in all experiments. Note that the ground truth images are not well aligned with RAW inputs, thus we adopt aligned PSNR, SSIM, and LPIPS as the evaluation metrics as in \cite{bhat2021deep}.

\noindent \textbf{Training details.}
We train the proposed KBNet on aforementioned synthetic datasets for 300 epochs. And as a common practice, we fine-tune the trained model on real-world dataset for 40 epochs. During training, the burst size is fixed to $N=8$ and the batch size is 16. We use Adam \cite{kingma2014adam} optimizer with $\beta_1$=0.9, $\beta_2$=0.999 and $\epsilon$=$10^{-8}$. The learning rate is initialized as 0.0002 and then decreases to half every 100 epochs. Our models are implemented by the PyTorch framework with 2 Titan Xp GPUs.

\subsection{Ablation Study}

In this section, We conduct ablation study to analyze the impact of the main components of the proposed framework: adaptive kernel-aware block (\textbf{AKAB}), kernel-aware deformable convolution (\textbf{KAD}) and \textbf{Pyramid KAD}. In addition, we also give an attention to the \textbf{RCAB} in reconstruction. To conveniently illustrate the superiority of each module, we implement a baseline model (KBNet-A) which only contains a restorer that adopts normal DCN as the alignment module and uses residual blocks for both feature extraction and HR image reconstruction. All methods are evaluated on both the synthetic data and the real data.

The comparison among the baseline and our methods with different modules (KBNet-B through KBNet-E) are reported in \Cref{table:ablation}, from which we have the following observations. First, the AKAB module plays a vital role for extracting useful features by considering multiple degradation information. Compared with the baseline, the model with AKAB improves the results about 1.3$+$ dB on the synthetic dataset and 0.4 dB on the real dataset. Second, the degradation and multi-scale information is also essential in the alignment. By utilizing the multi-scale features and kernel maps, the Pyramid KAD could achieve an impressive performance even the burst frames are noisy and blurry. Third, although the RCAB performs well on synthetic dataset, the improvement is incremental when finetune it on the real world dataset, which further demonstrates that the AKAB and the Pyramid KAD are key contributions of our work.

We also retrain an EDVR~\cite{wang2019edvr} model and our KBNet with the Pyramid, Cascading, and Deformable (PCD) alignment module~\cite{wang2019edvr} on the datasets. The results are shown in Table \ref{table:edvr}. With the proposed pyramid KAD, our model can perform alignment with degradation information in each pyramid level, which leads to important improvement.

\begin{table}[ht]
\setlength{\abovecaptionskip}{0.05in}
\setlength{\belowcaptionskip}{-0.1in}
\centering
\resizebox{1.\linewidth}{!}{
\begin{tabular}{lcccccc}
\toprule

\multirow{2}{*}{Method} & \multicolumn{3}{c}{Synthetic}  & \multicolumn{3}{c}{Real-world}     \\ \cmidrule(lr){2-4} \cmidrule(lr){5-7}
& PSNR$\uparrow$  & SSIM$\uparrow$   & LPIPS$\downarrow$    & PSNR$\uparrow$  & SSIM$\uparrow$   &LPIPS $\downarrow$  \\ \midrule

EDVR~\cite{wang2019edvr} & 36.34 &0.906 &0.138  & 47.48 &0.982 &0.031 \\
KBNet+PCD  & 37.02 &0.920 &0.124  & 48.14 &0.984 &0.026        \\
% \midrule
KBNet+Pyramid KAD  & \textbf{37.29} & \textbf{0.922} & \textbf{0.120}  & \textbf{48.27} &\textbf{0.986} &\textbf{0.025} \\

\bottomrule
\end{tabular}}
% }
\caption{Comparison of our method with EDVR.}
\label{table:edvr}
\end{table}

\begin{table*}[t]
\setlength{\abovecaptionskip}{0.05in}
\setlength{\belowcaptionskip}{-0.05in}
\centering
\resizebox{.98\linewidth}{!}{
\begin{tabular}{lccccccccc}
\toprule
% \hline
\multirow{2}{*}{Method} & \multicolumn{3}{c}{$\sigma=[0,1.6]$}              & \multicolumn{3}{c}{$\sigma=[1.6, 3.2]$}  & \multicolumn{3}{c}{$\sigma=[3.2, 4.8]$}   \\ \cmidrule(lr){2-4} \cmidrule(lr){5-7}  \cmidrule(lr){8-10}
& PSNR$\uparrow$  & SSIM$\uparrow$   & LPIPS$\downarrow$    & PSNR$\uparrow$  & SSIM$\uparrow$   & LPIPS$\downarrow$  & PSNR$\uparrow$  & SSIM$\uparrow$   & LPIPS$\downarrow$ \\ 

\midrule
        
DAN$^*$ \cite{luo2020unfolding} & 33.38 &0.8543 &0.1712        &32.69 & 0.8321 & 0.2210     &31.98 &0.8255 &0.2688  \\
    
DBSR \cite{bhat2021deep} & 35.52 &0.9086 &0.1300        &35.94 & 0.9015 & 0.1625     &33.34 &0.8633 &0.2730  \\
EBSR \cite{luo2021ebsr} & 35.67 &0.9156 &0.1149        &36.39 & 0.9095 & 0.1434     &33.57 &0.8673 &0.2669  \\
DeepREP \cite{bhat2021deepre} & 36.46 &0.9233 &0.1104        &36.26 & 0.9082 & 0.1510     &33.49 &0.8664 &0.2721  \\
    
KBNet(Ours) & \textbf{37.43} &\textbf{0.9314} &\textbf{0.0967}        &\textbf{37.27} & \textbf{0.9172} & \textbf{0.1237}     &\textbf{35.28} &\textbf{0.8924} &\textbf{0.1941}  \\

% \hline
\bottomrule
\end{tabular}
}
\captionof{table}{Comparison of our method with existing MFSR approaches on the Synthetic test dataset, for scale factor 4. The kernel width $\sigma$ is split into three ranges. `*' means it is a single image blind super-resolution method.}
\label{table:syn}
\end{table*}

\begin{figure*}
\centering
	\begin{minipage}[h]{.97\linewidth}
	\setlength{\abovecaptionskip}{0.0in}
    \setlength{\belowcaptionskip}{0.1in}
		\centering
		\includegraphics[width=.95\linewidth]{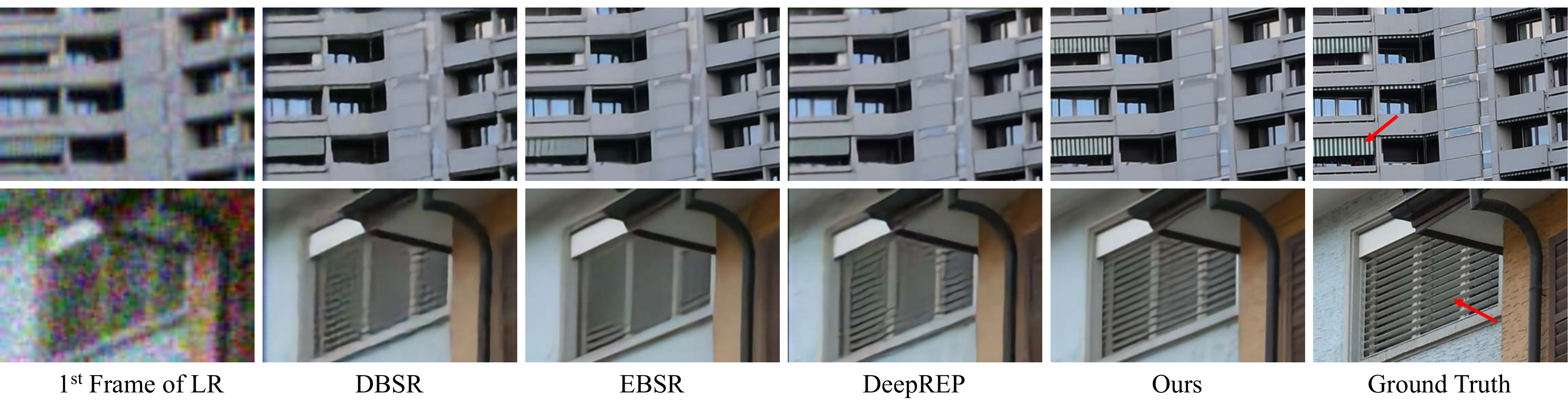}
		\caption{Qualitative comparison of our method with other MFSR approaches on \textbf{synthetic} dataset.}
		\label{fig:cmp_syn}
	\end{minipage}
	\begin{minipage}[h]{.97\linewidth}
	\setlength{\abovecaptionskip}{0.0in}
    \setlength{\belowcaptionskip}{-0.0in}
		\centering
		\includegraphics[width=.95\linewidth]{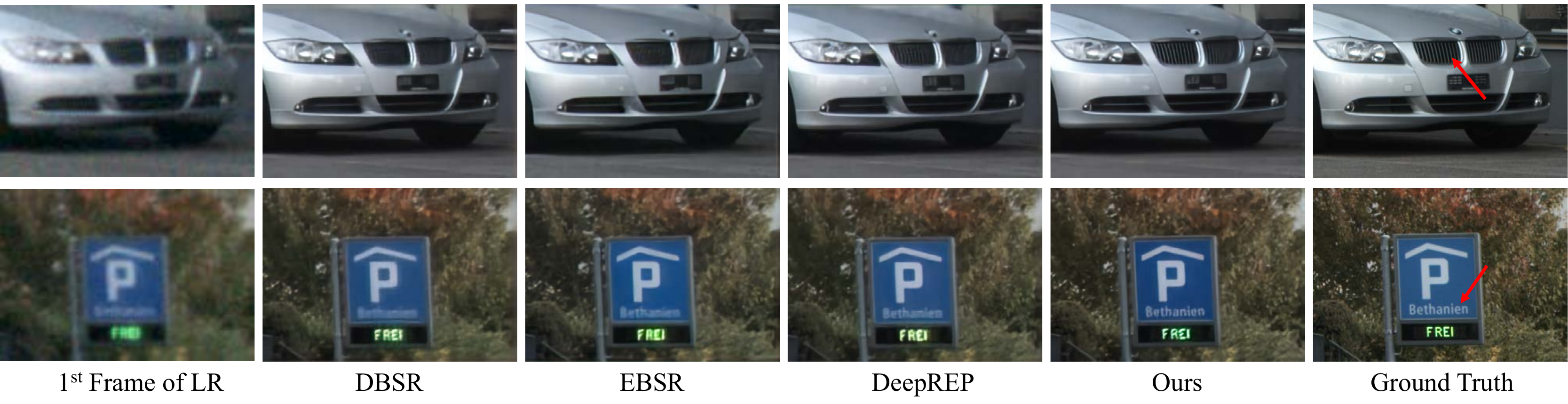}
		\caption{Qualitative comparison of our method with other MFSR approaches on real-world \textbf{BurstSR} dataset.}
		\label{fig:cmp_real}
	\end{minipage}
\end{figure*}

\subsection{Comparisons with State-of-the-Art Methods}

We compare KBNet with other state-of-the-art learning-based raw bursts SR methods, such as DBSR \cite{bhat2021deep}, EBSR \cite{luo2021ebsr}, and DeepREP \cite{bhat2021deepre}. Both DBSR and DeepREP are proposed by Bhat et al. \cite{bhat2021deep,bhat2021deepre}. The former uses a flow-based alignment network with an attention-based fusion to handle the raw burst inputs. The latter employs a deep reparametrization of the MAP to solve image restoration problems. And EBSR is the winner solution of the NTIRE21 Burst Super-Resolution Challenge \cite{bhat2021ntire}. All of these methods are implemented from their official code repositories and re-trained with our multiple degradation setting following Sec. \ref{ex:setting}. We also finetune these models and evaluate them on the real-world dataset. The burst sizes of all methods are fixed to 14. In addition, we compare a single image blind SR model DAN \cite{luo2020unfolding} which estimates a kernel for the first burst frame and restores the HR image conditioned on that kernel and the first LR frame. Note that DAN adopts an iteratively predicting strategy and chooses to stretch the kernel embedding in reconstruction as the same as IKC \cite{gu2019blind}.

% \begin{figure*}[ht]
% \setlength{\abovecaptionskip}{0.in}
% \setlength{\belowcaptionskip}{-0.1in}
%   \centering
%   \includegraphics[width=1.\linewidth]{figs/cmp_syn.pdf}
%   \caption{Visual comparison of our approach with other methods on synthetic dataset.}
%   \label{fig:cmp_syn}
% \end{figure*}

% % \vspace{-0.1in}

% \begin{figure*}[ht]
% \setlength{\abovecaptionskip}{0.in}
% \setlength{\belowcaptionskip}{-0.1in}
%   \centering
%   \includegraphics[width=1.\linewidth]{figs/cmp_real.pdf}
%   \caption{Visual comparison of our approach with other methods on real images.}
%   \label{fig:cmp_real}
% \end{figure*}

% \newpage
% \vspace{0.1in}

\noindent \textbf{Evaluation on synthetic data.}
Firstly, we evaluate the proposed KBNet on the synthetic dataset as introduced in \cref{ex:setting}. Quantitative results are shown in \Cref{table:syn}. Our method achieves the best results and significantly outperforms other burst super-resolution methods. 
% Since each RAW frame could be blurred with an anisotropic Gaussian kernel, the single image method DAN \cite{luo2020unfolding} can not restore a pleasant SR result based on that single blurry input. While the MFSR methods can utilize the complementary information of different frames to obtain rich details. 
As the table illustrated, all the MFSR methods outperform DAN \cite{luo2020unfolding} with great improvements of 3+ dB on kernels width in range $[1.6, 3.2]$ in terms of PSNR. These MFSR mthods do not explicitly utilize degradation information in the restoring and thus are powerless when facing complex degradations. In contrast, the proposed KBNet significantly outperforms other MFSR methods over all kernel width ranges. The qualitative comparison are shown in Fig. \ref{fig:cmp_syn}. 
The super-resolved images produced by our KBNet are visually pleasant and have rich details, which demonstrates the superior of our method, and indicates that involving degradation information into restoration can help to obtain informative features and thus improve SR results.

\noindent \textbf{Evaluation on real-world data.}
Now we conduct the experiment of evaluating models that are pre-trained on synthetic dataset and finetuned on the real-world dataset. Note that the ground-truth kernels of real-world images are not available, which are required by the KBNet in the kernel estimation learning process. Alteratively, we freeze the kernel estimator and only finetune the image restorer. The quantitative results are shown in \Cref{table:real}. As we can see, the DeepREP \cite{bhat2021deepre} significantly outperforms DBSR \cite{bhat2021deep} and EBSR \cite{luo2021ebsr}, but is still inferior to the proposed KBNet. Visual comparisons on the real-world images are shown in Fig. \ref{fig:cmp_real}. It is obviously that the results produced by the KBNet have favorable perceptual quality in edges and details, and is robust to real-world noises.

\subsection{Analysis about kernels and frame numbers.}

\begin{figure}[t]
\setlength{\abovecaptionskip}{0.05in}
  \centering
  \includegraphics[width=1.\linewidth]{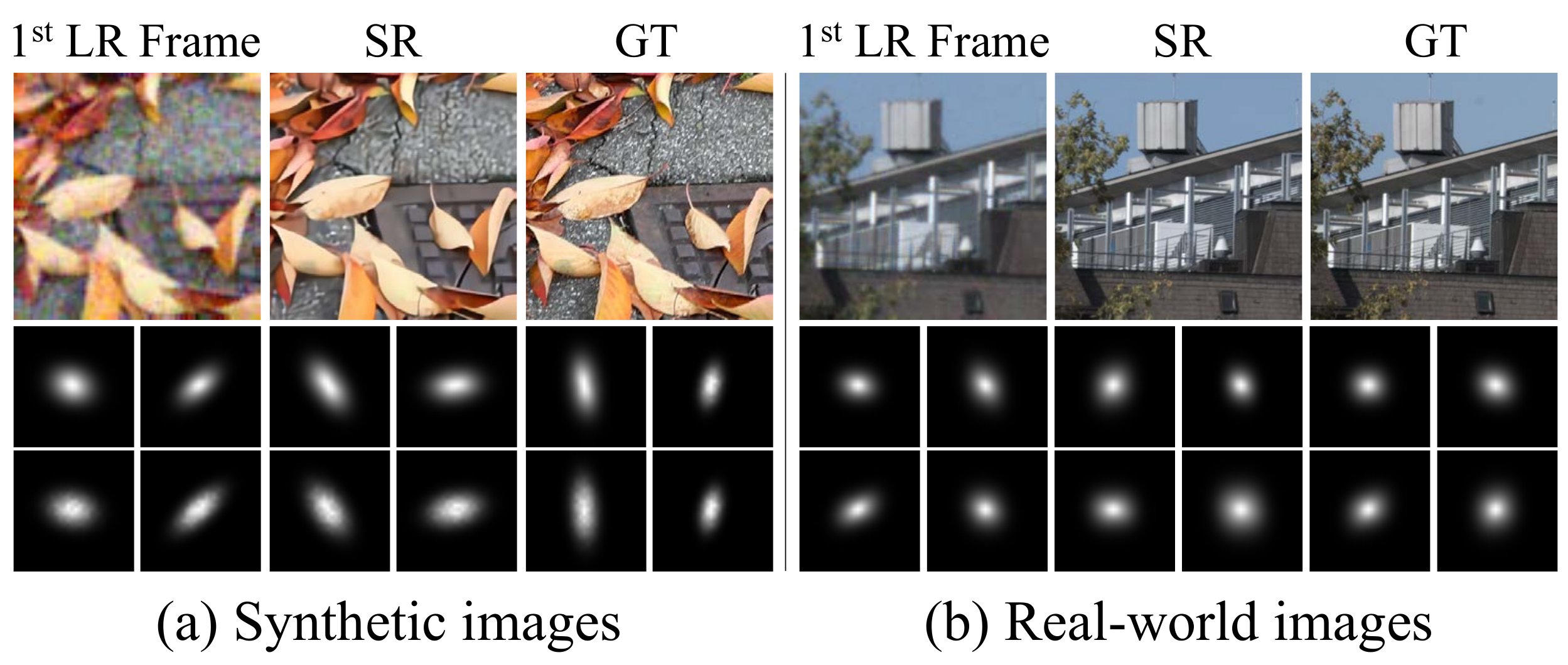}
  \caption{Examples of estimated kernels and super-resolved images. \textbf{(a)} The first 6 kernels of the corresponding synthetic image. The top row of the kernel is the estimated kernels and second row of kernel is the ground-truth kernels. \textbf{(b)} The first 12 kernels of the corresponding real image.}
  \label{fig:show_k}
\end{figure}

We assumes that all frames of a burst sequence are captured by a smartphone under the burst shooting mode in which different degradation kernels can be produced by hand-shaking or different shooting parameters. Visual examples of the estimated kernels are shown in Fig. \ref{fig:show_k}.
On the synthetic setting, most kernels can be accurately estimated by our method, which helps us to restore HR images. For real-world images, the degradation kernel should always follow the Gaussian distribution depends on the depth at each pixel and the focal length of the camera~\cite{chaudhuri1999depth}. Thus our method prefers to generate some Gaussian-like kernels despite the kernel estimator is not optimized to fit the real dataset. The experiment in Table \ref{table:real-kernel} illustrates that the estimated kernels are also useful for restoring HR images.

Moreover, we investigate the impact of multiple frames and compare KBNet with other MFSR methods. Here we don't conduct the results of EBSR because both it's training and testing require fixing the frame number to 14. The results on PSNR are shown in Fig. \ref{fig:cmp_psnr}. As the frame number increases, all MFSR methods could achieve higher performances. The proposed KBNet outperforms other methods over all frames and datasets.

\begin{table}[t]
\setlength{\abovecaptionskip}{0.05in}
\centering
\resizebox{1.\linewidth}{!}{
\setlength{\tabcolsep}{1.mm}{
\begin{tabular}{lccccc}
\toprule

Method & DAN$^*$ \cite{luo2020unfolding}  & DBSR \cite{bhat2021deep}    & EBSR \cite{luo2021ebsr}  & DeepREP \cite{bhat2021deepre}    & KBNet(Ours)     \\  \hline

PSNR$\uparrow$ & 46.18 &47.48 &47.25   & 48.15 & \textbf{48.27}   \\
SSIM$\uparrow$ & 0.9777 &0.9824 &0.9800   & 0.9842 & \textbf{0.9856}   \\
LPIPS$\downarrow$ & 0.0389 &0.0326 &0.0356   & 0.0265 & \textbf{0.0248}   \\

\bottomrule
\end{tabular}}
}
\caption{Quantitative comparison of the proposed method with existing MFSR approaches on the real-world BurstSR $\times$4 dataset. `*' means it is a single image blind SR method.}
\label{table:real}
\end{table}

\begin{table}[t]
\setlength{\abovecaptionskip}{0.05in}
\centering
\resizebox{.95\linewidth}{!}{
\begin{tabular}{lccc}
\toprule
Method & PSNR $\uparrow$  & SSIM $\uparrow$   & LPIPS $\downarrow$  \\ \hline
KBNet \textit{w/o} using kernel & 47.54 &0.9824 &0.0299          \\
KBNet, fixed Gaussian kernel & 48.12 &0.9838 &0.0264          \\
KBNet, estimated kernel & 48.27 &0.9856 &0.0248          \\

\bottomrule
\end{tabular}
}
\caption{Different kernel strategies on real-world dataset.}
\label{table:real-kernel}
\end{table}

% We also conduct ablation studies on synthetic datasets to analyze the impact of two main components in the proposed framework: adaptive kernel-aware block (AKAB) and kernel-aware deformable (KAD) alignment module. The baseline is a multi-frame SR method that adopts normal deformable convolution to align frames and reconstructs SR results by several residual blocks The results are reported in \Cref{table:ablation}. It is obvious that we can benefit a lot from introducing the degradation kernel. By using the AKABs, we obtain an improvement of 1.29 dB in PSNR. And the KAD further improves the result.

% \begin{table}[t]
% \centering
% \resizebox{.8\linewidth}{!}{
% \begin{tabular}{l|cccc}
% \toprule
% % \multirow{3}{*}{Method}     & \\
% % \multicolumn{9}{c}{Synthetic data}     \\ %\cline{2-10}

% Method & DBSR  & DeepREP & EBSR & KBNet \\ \hline

% \# Params & 13.011M \\ &12.131M & 76.765M & 20.898M \\
% Flops & 56.039G &70.981G    &191.539G &95.761G \\

% \bottomrule
% \end{tabular}
% }
% \caption{Comparison of the number of parameters and Flops}
% \label{table:param}
% \end{table}

% %%%%%%%%%%%%%%%%%%%%%%%%%%%

\subsection{Synthetic Model Transferring}

The motivation of this experiment is that making paired SR datasets for real-world photography applications is really challenging since the LR images and GT are usually captured from different devices (e.g., smartphones and DSLR cameras). The mismatch of image qualities and colors would make it extremely difficult to train a SR model across modalities. 
% Moreover, the obtained paired images often exist severe color mismatch which may deteriorate the training. In such a situation, DBSR \cite{bhat2021deep} propose to incorporate a pre-trained flow-based alignment network ( e.g., PWC-Net) and a global color mapping function to handle the spatial mis-alignment and color mis-match issues. However, the pre-trained PWC-Net still can not handle complicated real-world scenes. And incorporating that model in training is time-consuming and computationally costly. 
In such a situation, we would like to train the model only in the multi-degradation environment and apply it to the real scenes directly, which seems like a zero-shot transferring problem.

\begin{figure}[t]
\setlength{\abovecaptionskip}{0.04in}
\setlength{\belowcaptionskip}{-0.1in}
\centering
	\begin{minipage}[t]{0.495\linewidth}
		\centering
		\includegraphics[width=1.\linewidth]{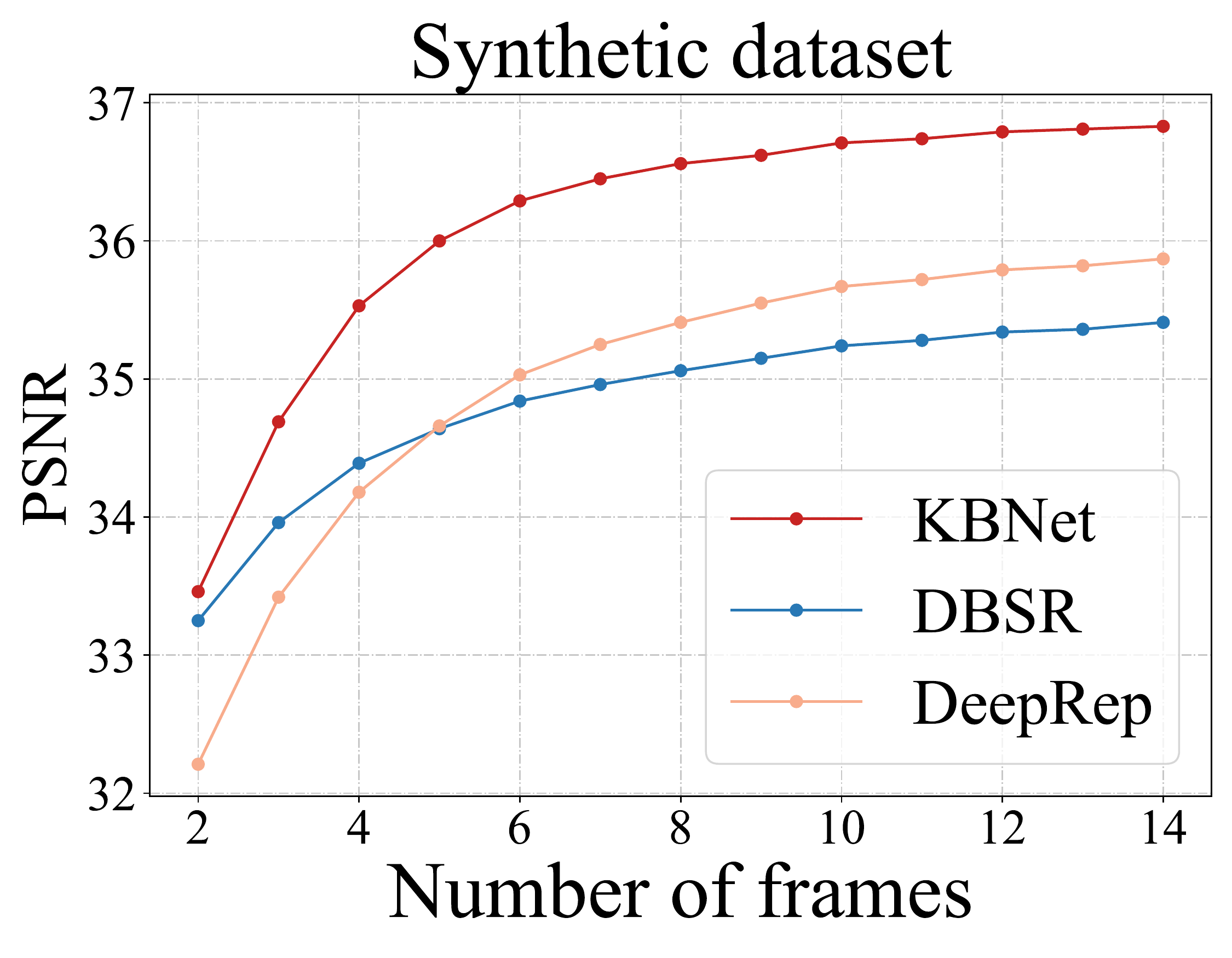}
	\end{minipage}
	\begin{minipage}[t]{0.495\linewidth}
		\centering
		\includegraphics[width=1.\linewidth]{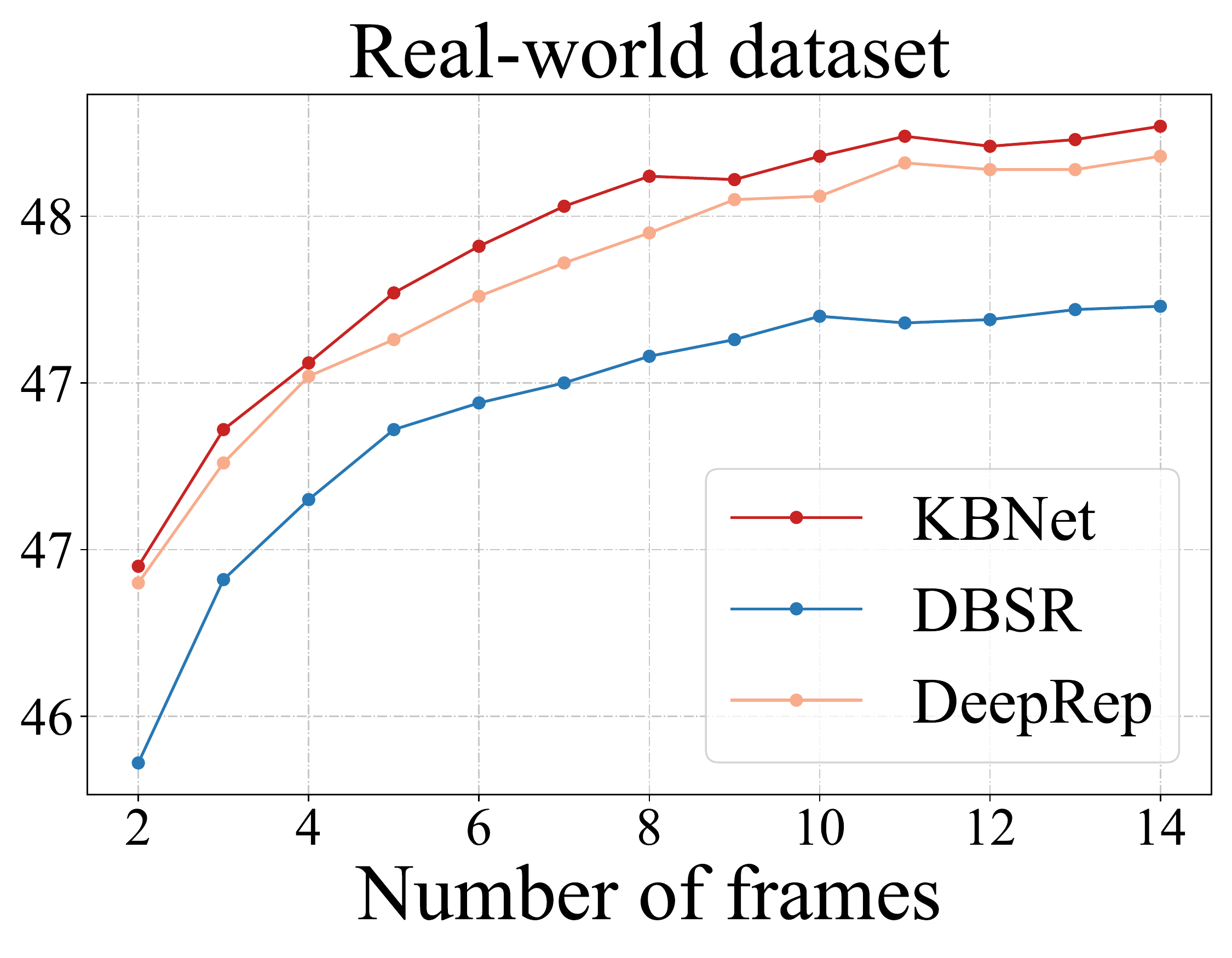}
	\end{minipage}
	\caption{Comparison of the PSNR performances among different MFSR approaches on synthetic and real-world datasets with different number of frames.}
	\label{fig:cmp_psnr}
\end{figure}

To illustrate the idea, we provide the comparison of transferring models under bicubic degradation and blind degradation in \Cref{table:transfer}. Under the multi degradation environment, all MFSR methods can achieve higher performances compared with their bicubicly trained models. And the proposed KBNet remarkablely outperforms other methods and can produce visually pleasant results on real images, which indicate that even if the KBNet is trained on synthesized image pairs, it still has the ability to generalize to images in real applications.

\begin{table}[t]
\setlength{\abovecaptionskip}{0.05in}
\setlength{\belowcaptionskip}{-0.1in}
\centering
\resizebox{1.\linewidth}{!}{
\setlength{\tabcolsep}{1.mm}{
\begin{tabular}{lcccccc}
\toprule

\multirow{2}{*}{Method} & \multicolumn{3}{c}{Bicubic Degradation}  & \multicolumn{3}{c}{Blind Degradation}     \\ \cmidrule(lr){2-4} \cmidrule(lr){5-7}
& PSNR$\uparrow$  & SSIM$\uparrow$   & LPIPS$\downarrow$    & PSNR$\uparrow$  & SSIM$\uparrow$   &LPIPS $\downarrow$  \\ \midrule

DBSR \cite{bhat2021deep} & 44.64 &0.967 &0.079 & 45.18(\textcolor{red}{+0.53}) &0.974(\textcolor{red}{+0.007}) &0.048(\textcolor{cyan}{-0.031}) \\
EBSR \cite{luo2021ebsr} & 44.32 &0.963 &0.082  & 44.96(\textcolor{red}{+0.64}) &0.970(\textcolor{red}{+0.007}) &0.048(\textcolor{cyan}{-0.034})        \\
DeepREP \cite{bhat2021deepre} & 44.80 &0.968 &0.080   & 45.39(\textcolor{red}{+0.59}) &0.974(\textcolor{red}{+0.006}) &0.045(\textcolor{cyan}{-0.035})   \\
\midrule
KBNet(Ours) & - & - & -  & \textbf{45.68} &\textbf{0.979} &\textbf{0.042} \\

\bottomrule
\end{tabular}}
}
\caption{Quantitative results of transferring models from different synthetic environments  to real-world images. The improvements on each metric are marked in colors.}
\label{table:transfer}
\end{table}

\section{Conclusion}
In this paper, we present a new framework, named KBNet, to handle the multi-frame super-resolution problem with considering multiple complicated degradations. The proposed burst blind super-resolution task is highly related to real-world applications. To address it, we introduce a kernel-based multi-frame restoration network that includes an adaptive kernel-aware block (AKAB) and a pyramid kernel-aware deformable (Pyramid KAD) alignment module. The blur kernels are first estimated by an estimator and then fed to the LR feature extraction module as well as the feature alignment module to generate a super-resolved clear image. The proposed method can be end-to-end trained on the synthetic dataset and evaluated on both synthetic and real-world images. Experiment results demonstrate that our method can achieve a good performance on various degradations and is beneficial to real-world device applications.

\noindent \textbf{Acknowledgments}
Thanks to Yanqing Lian, Youqin Zheng, Wenjing Lian, Jingyuan Lian, and Ziwei Luo for their selfless support and help in this work. We also thank the anonymous reviewers for helping improve our work.

{\small
\bibliographystyle{ieee_fullname}
\bibliography{egbib}
}

\end{document}